\documentclass[letterpaper, 10 pt, conference]{ieeeconf}  

\IEEEoverridecommandlockouts                              

\long\def\devour#1{}

\newcounter{notecounter}
\newcommand{\enotesoff}{\long\gdef\enote##1##2{}}

\enotesoff

\usepackage{url}
\usepackage{xcolor}
\usepackage{booktabs}
\usepackage{multirow}
\usepackage{graphicx}

\title{\LARGE \bf
LoHoRavens: A Long-Horizon Language-Conditioned Benchmark for Robotic Tabletop Manipulation
}

\author{Shengqiang Zhang~$^{1, 3}$, Philipp Wicke~$^{1, 3}$, Lütfi Kerem Şenel~$^{1, 3}$, \\ Luis Figueredo~$^{2}$, Abdeldjallil Naceri~$^{2}$, Sami Haddadin~$^{2}$, Barbara Plank~$^{1, 3}$, Hinrich Schütze~$^{1, 3}$ \\
$^{1}$~CIS, LMU Munich \quad
$^{2}$~RSI, MIRMI, TUM \\
$^{3}$~Munich Center for Machine Learning (MCML)
}

\begin{document}

\maketitle
\thispagestyle{empty}
\pagestyle{empty}

\begin{abstract}

The convergence of embodied agents and large language models (LLMs) has brought significant advancements to embodied instruction following.
Particularly, the strong reasoning capabilities of LLMs make it possible for robots to perform long-horizon tasks without expensive annotated demonstrations.
However, public benchmarks for testing the long-horizon reasoning capabilities of language-conditioned robots in various scenarios are still missing. 
To fill this gap, this work focuses on the tabletop
manipulation task and releases a simulation benchmark,
\textit{LoHoRavens}, which covers various long-horizon
reasoning aspects spanning color, size, space, arithmetics
and reference.
Furthermore, there is a key modality bridging problem for
long-horizon manipulation tasks with LLMs: how to
incorporate the observation feedback during robot execution
for the LLM's closed-loop planning, which is however less studied by prior work. 
We investigate two methods of bridging the modality gap: caption generation and learnable interface for incorporating explicit and implicit observation feedback to the LLM, respectively.
These methods serve as the two baselines for our proposed benchmark. 
Experiments show that both methods struggle to solve some tasks, indicating long-horizon manipulation tasks are still challenging for current popular models.
We expect the proposed public benchmark and baselines can help the community develop better models for long-horizon tabletop manipulation tasks.\footnote{The video and code of LoHoRavens are available at~\url{https://cisnlp.github.io/lohoravens-webpage/}.}

\end{abstract}

\section{INTRODUCTION}
In embodied instruction following, an embodied agent such as
a robot
is given a language based instruction and expected to
follow the instruction to complete the designated task.
Of particular interest is long-horizon instruction following:
how to endow embodied agents with long-horizon instruction
following capabilities attracts more and more attention,
because it is more in line with the daily life scenes
that are of practical importance in robotics.
The long-horizon task usually includes a quite high-level
instruction and cannot be completed in just a few
steps. Thus, the embodied agent must understand the language instruction well and perform long-horizon memorizing and complex reasoning.
Thanks to the emergent abilities of large language models
(LLMs)~\cite{wei2022emergent}, embodied agents are able to
borrow the rich knowledge and commonsense about the world
and the strong reasoning capabilities from LLMs, reducing
the need for large expensive datasets of expert annotated demonstrations.
With LLMs, embodied agents show better and better impressive performance on long-horizon tasks~\cite{saycan2022arxiv, driess2023palme, brohan2022rt1, brohan2023rt}.

\begin{figure}
    \centering
    \includegraphics[width=.45\textwidth]{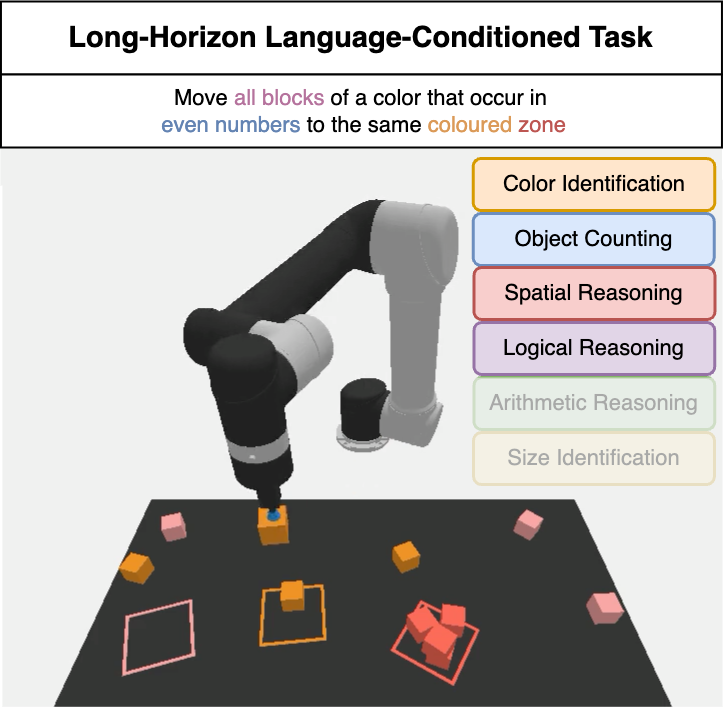}
    \caption{A long-horizon task such as ``Move all blocks
      of a color that occur in even numbers to the same
      coloured zone'' requires various different reasoning
      capabilities that go beyond a simple pick-and-place
      task. In this example, the instruction requires the
      model to identify colors (red, pink and orange), count
      objects (4x orange, 4x red, 3x pink), identify spatial
      components (orange area, pink area, red area) and
      understand the logic behind the task: select either
      orange or red as the color (only the number of
      orange/red blocks is even) and then move the blocks of
      the selected color into the zone of that color.}
    \label{fig:enter-label}
\end{figure}

\begin{figure*}[t]
    \centering
    \includegraphics[width=\textwidth]{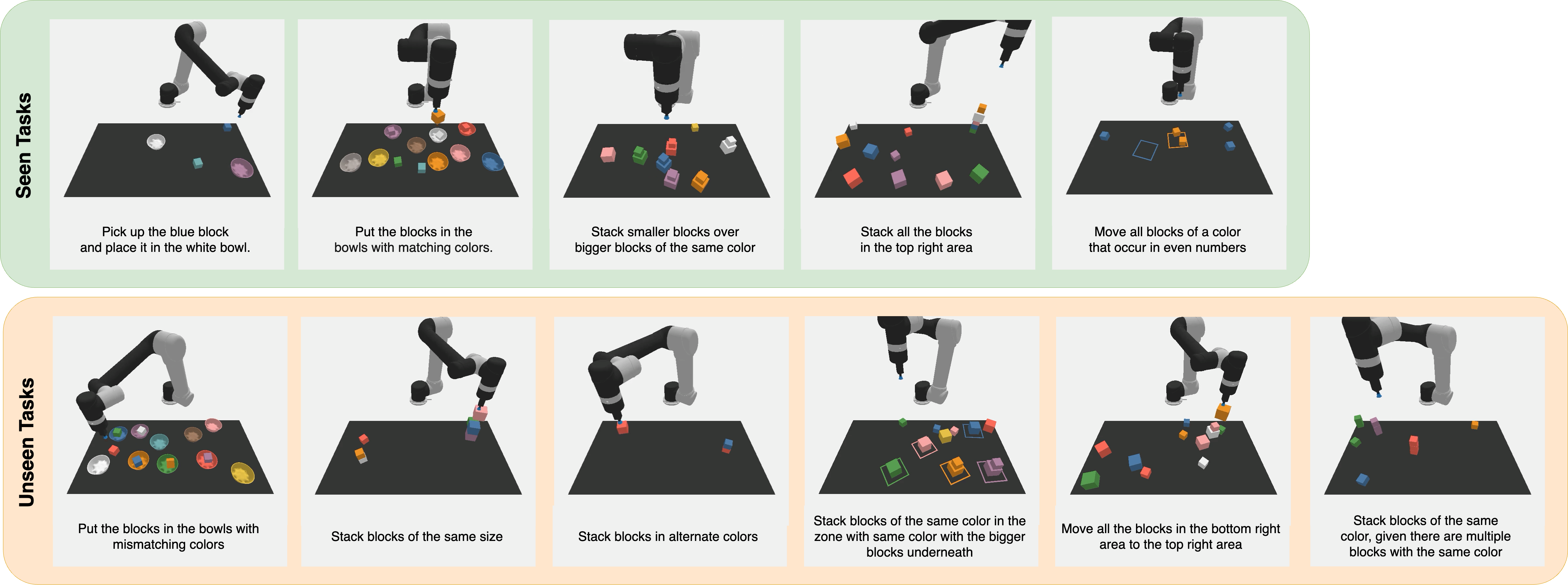}
    \caption{Example screenshots of the five seen and six unseen LoHoRavens tasks.}
    \label{fig:all_figures}
\end{figure*}

This work focuses on language-conditioned robotic tabletop manipulation tasks.
To develop better robots for long-horizon manipulation tasks, good benchmarks are essential to test their capabilities.
However, most current benchmarks either do not focus on long-horizon tasks or are not language-conditioned. 
Meta-World~\cite{yu2020meta} is a simulated robotic manipulation benchmark for meta-reinforcement learning and multi-task learning, but its tasks  are neither language-conditioned nor long-horizon.
RLBench~\cite{james2020rlbench} introduces 100 simulated household tasks with corresponding natural language instructions;
Ravens~\cite{zeng2021transporter, shridhar2022cliport}, Robosuite~\cite{robosuite2020}, and VIMA-Bench~\cite{jiang2023vima} introduce various language-conditioned tabletop manipulation tasks with robot arms.
However, these four benchmarks do not focus on long-horizon tasks.
FurnitureBench~\cite{heo2023furniturebench} 
and CausalWorld~\cite{ahmed2020causalworld}
focus on
real-world furniture assembly and 3D shape construction respectively, both of which are complex and
long-horizon manipulation tasks; but they are not language-conditioned benchmarks.
CALVIN~\cite{mees2022calvin} is a long-horizon
language-conditioned public benchmark, but step-by-step
instructions are provided to complete each long-horizon
task, without the need for
any long-horizon reasoning by the robot.
Inner Monologue~\cite{pmlr-v205-huang23c} and
CoP~\cite{codeaspolicies2022} have experiments on
long-horizon language-conditioned manipulation tasks. However, they do not open-source their simulated environments and tasks.
Language-Table~\cite{lynch2022language-tabel} is a multitask
language-labeled continuous control benchmark with
long-horizon goal tasks included. Unfortunately, the parts for long-horizon tasks are not released.


To fill this gap and benefit the open-source community, we develop a long-horizon language-conditioned simulated benchmark, called~\textbf{LoHoRavens}, for robotic tabletop manipulation tasks and open-source it.
LoHoRavens is built based on the Ravens robot simulator and contains ten long-horizon language-conditioned tasks in total.
The tasks are split into seen tasks and unseen tasks to evaluate the robot's generalization performance.
We define tasks in which  the robot needs to execute at least five pick-and-place steps to complete the high-level instruction as a long-horizon task.
LoHoRavens first requires the robot agent to understand the deep semantics of each high-level instruction well.
Then LoHoRavens covers various long-horizon reasoning
aspects including color, size, space, arithmetics and reference.
To solve each task, the robot must combine several of the
reasoning capabilities and
develop its long-horizon plan accordingly.
Following previous work~\cite{pmlr-v205-huang23c, guo2023doremi}, LoHoRavens also further boosts the complexity of each task by perturbing the environment to increase the probability of execution failure, such that the robot has to incorporate real-time observation feedback for the long-horizon planning.

Fig.\ \ref{fig:enter-label} gives an example of 
a long-horizon task that
requires reasoning
capabilities that go beyond a simple pick-and-place
task. 
Fig.\ \ref{fig:all_figures} gives
example screenshots of the
eleven tasks of the LoHoRavens benchmark:
five seen tasks and six unseen tasks.


To solve the challenging LoHoRavens benchmark tasks, a key
modality bridging problem arises: although using LLMs as
planners has been a popular method in robotics, how to
incorporate the observation feedback during the robot's
execution for the LLM's closed-loop long-horizon planning is
still an under-explored problem.


In this work, we investigate two methods for modality
bridging:
the \emph{explicit method} of
caption generation and the \emph{implict method} of
learnable interface.
Explicit/implicit here refers to whether the observation
feedback is given in the form of explicit (human-readable)
natural language or in the form of an implicit
(non-human-readable) representation of the observation feedback.
These two methods will serve as strong baselines
for our proposed LoHoRavens benchmark.

The caption
generation method is shown in Fig.\ \ref{fig:architecture1}.
It uses a vision-language model (VLM) with
few-shot prompting to generate the descriptions of the
observation and the robot's execution states as the
(explicit) language feedback for the LLM's closed-loop
planning.

The learnable interface
method is shown in Fig.\ \ref{fig:architecture2}.
It trains a
multi-layer perceptron (MLP) to translate visual embeddings
of the observation to token embeddings that can be accepted
by LLMs as the (implicit) feedback for the LLM's closed-loop
planning.  


The extensive experiments on LoHoRavens benchmark show that
the proposed two baselines have a strong positive impact on  long-horizon
manipulation task performance.  But
both methods still struggle to solve
most of the long-horizon tasks. For tasks requiring reference
resolution,
we conjecture that
further
strategies need to be used to improve the LLM's
reference capabilities.  Overall the experimental results
indicate that the long-horizon language-conditioned
manipulation tasks are still challenging for current popular
models.  We hope our LoHoRavens benchmark
and the two baselines can help with developing more advanced
robots.


\section{LoHoRavens BENCHMARK}

\begin{table*}[]
\centering
\caption{LoHoRavens benchmark tasks and the experimental results of the two baselines. }
\label{tab:lohoravens}
\begin{tabular}{@{}cll|rrr|r@{}}
\toprule
\multicolumn{3}{c|}{\multirow{3}{*}{LoHoRavens Tasks}}                                                                                          & \multicolumn{3}{c|}{Explicit feedback}                                             & Implicit feedback      \\
\multicolumn{3}{c|}{}
& CLIPort               & \multirow{2}{*}{+Llama 2} &
  +Open
& \multirow{2}{*}{LLaVA} \\
\multicolumn{3}{c|}{}                                                                                                                           & (oracle)              &                           &       Flamingo                         &                        \\ \midrule
\multicolumn{1}{c|}{\multirow{6}{*}{\begin{tabular}{c}Seen\\ tasks\end{tabular}}}   & A.                  & Pick-and-Place primitives                                            & 61.2                  & 67.3                      & 67.3                           & 67.3                   \\
\multicolumn{1}{c|}{}                              & B.                  & ``Put the blocks in the bowls with matching colors"                  & 19.7                  & 27.9                      & 31.4                           & 37.0                   \\
\multicolumn{1}{c|}{}                              & C.                  & ``Stack smaller blocks over bigger blocks of the same color"         & 12.1                  & 17.5                      & 18.0                           & 22.1                   \\
\multicolumn{1}{c|}{}                              & D.                  & ``Stack all the blocks in the [ABS\_POS] area"                       & 22.5                  & 28.8                      & 30.4                           & 35.8                   \\
\multicolumn{1}{c|}{}                              & \multirow{2}{*}{E.} & ``Move all blocks of a color that occur in even numbers              & \multirow{2}{*}{13.4} & \multirow{2}{*}{9.1}      & \multirow{2}{*}{9.6}           & \multirow{2}{*}{8.2}   \\
\multicolumn{1}{c|}{}                              &                     & \ \ to the same colored zone"                                        &                       &                           &                                &                        \\ \midrule
\multicolumn{1}{c|}{\multirow{8}{*}{\begin{tabular}{c}Unseen\\ tasks\end{tabular}}} & F.                  & ``Put the blocks in the bowls with mismatching colors"               & 17.3                  & 24.8                      & 28.5                           & 21.1                   \\
\multicolumn{1}{c|}{}                              & G.                  & ``Stack blocks of the same size"                                     & 2.1                   & 15.8                      & 21.9                           & 14.7                   \\
\multicolumn{1}{c|}{}                              & H.                  & ``Stack blocks in alternate colors"                                  & 1.8                   & 8.7                       & 13.2                           & 5.2                    \\
\multicolumn{1}{c|}{}                              & \multirow{2}{*}{I.} & ``Stack blocks of the same color in the zone with same color,        & \multirow{2}{*}{8.5}  & \multirow{2}{*}{13.6}     & \multirow{2}{*}{12.8}          & \multirow{2}{*}{11.7}  \\
\multicolumn{1}{c|}{}                              &                     & \ \ with the bigger blocks underneath"                               &                       &                           &                                &                        \\
\multicolumn{1}{c|}{}                              & J.                  & ``Move all the blocks in the [ABS\_POS] area to the [ABS\_POS] area" & 15.1                  & 19.7                      & 27.4                           & 27.2                   \\
\multicolumn{1}{c|}{}                              & K.                  & ``Stack blocks of the same color"                                     & 6.7                   & 3.5                      & 4.0                           & 6.8                   \\ \bottomrule
\end{tabular}%
\end{table*}

As far as we know, LoHoRavens is the first public benchmark for long-horizon language-conditioned robotic tabletop manipulation tasks without giving step-by-step instructions for the high-level goal of each task.
In this section, we give details about
the composition of the benchmark,
how we build it and how we evaluate
long-horizon language-conditioned systems on the benchmark.

\subsection{Simulation environment}
\textbf{LoHoRavens} is built based on the \textbf{Ravens} robot simulator by extending it to \textbf{Lo}ng-\textbf{Ho}rizon tasks.
In the LoHoRavens simulation environment, there are a UR5e robot arm with a suction gripper and some objects on the table. Given a high-level language based instruction (e.g., ``stack all the blocks of the same size"), the robot is supposed to rearrange these objects to a desired state.
Based on Ravens, the observation space includes an RGB-D reconstruction from three camera views (front, left and right view). Besides, we also provide an RGB image rendered from the top-down view to the observation space.
The action space of LoHoRavens consists of a language-conditioned pick-and-place motion primitive which is parameterized by object names.

Currently, LoHoRavens contains ten long-horizon tasks in total (see Table~\ref{tab:lohoravens}).
To support more complex long-horizon reasoning, besides the
vanilla pick-and-place primitive (e.g., ``pick up the red
block and place it on the yellow block"), we add two other
pick-and-place primitives.\footnote{
We use the expression ``pick-and-place primitives" to
refer to all three primitives in the table.} One is related to size
reasoning (e.g., ``pick up the smaller red block and place it on the bigger yellow block"), the other is related to spatial reasoning (e.g., ``pick up the red block and place it in the top right area").
In addition to the pick-and-place primitive,
we borrow two interesting tasks (tasks B and F) from Inner Monologue and CoP, and design another eight long-horizon tasks by ourselves. 

Unlike Ravens and VIMA-Bench's complicated and various objects, LoHoRavens only contains three kinds of objects: block, bowl, and zone (see Fig. \ref{fig:all_figures})
because we do not want to test the robot's generalization
capability to new or unseen objects in this work. Instead,
we focus  on the long-horizon reasoning capabilities which
are related to the general attributes of objects like size,
color and spatial position. Such reasoning capabilities can be generalized to other objects as well.
In addition to these general object attributes, we are also interested in the reasoning capabilities related to attributes of multiple objects. So we include several tasks to test arithmetic and reference reasoning capabilities (e.g., tasks E and K).

To simulate the disturbance in the real world, we add noises and perturbations to the robot's environment at test time. 
Following Inner Monologue~\cite{pmlr-v205-huang23c}, we add Gaussian noise $\mathcal{N}(0, 3)$ for pixel observations and $\mathcal{N}(0, 2.5)$ for policy primitives. Moreover, we add a dropping probability $p$ for the end-effector to drop the picked block every second following DoReMI~\cite{guo2023doremi}.

\subsection{Dataset}
Like Ravens and VIMA-Bench, our simulator can also generate expert demonstrations automatically with the scripted oracle program. 
The oracle agent has access to the ground-truth pick and place poses and uses pre-specified heuristics to complete the tasks.
All the tasks can be instantiated into thousands of task instances with different random seeds.
To train the pick-and-place primitives,  we  generate 20,000 demonstrations for each primitive.
To build the benchmark, we generate 1,000 demonstrations
as the train set, 200 demonstrations as the validation
set, and 200 demonstrations as the test set for each long-horizon
task. Note that the colors of objects
are chosen randomly, so they are generally different
in training, validation and test sets.
We split all the tasks into seen tasks and unseen tasks. The seen tasks are used for training and writing prompts. The unseen tasks are used for evaluating the model's generalization abilities to new tasks.
Most of the task instances need five or more steps to complete. However, due to the attributes of some tasks, it is difficult to design a high-level goal that needs many steps. Taking stacking blocks as an example, it is difficult to stack more than five blocks in the same position because the blocks will easily fall if they are stacked too high.

\subsection{Evaluation}
Depending on the task, there are two different match methods for evaluating whether the states of the objects are correct compared to the ground-truth states. One is based on ``pose match'', which means an object's position and rotation should be the same as the ground-truth one. Another one is based on ``zone match'', which means the overlap area of two objects should be larger than a threshold.
Following Ravens and CLIPort, LoHoRavens adopts a score from 0 (fail) to 100 (success) to evaluate the final state for each task instance.
The score assigns the partial rewards according to the total
number of pick-and-place steps for each task instance. For
example, if a task needs ten pick-and-place steps to
complete, and the test model finishes eight of them, the score for this instance would be $8 / 10 = 80 \%$.

\section{BASELINES}

\begin{figure*}[]
  \caption{\textbf{Explicit feedback: Caption generation.}
    This baseline takes the human input ("Move all blocks of a color that occur in even numbers to the same coloured zone") and asks an LLM to create the next step that needs to be done in order to achieve the task. The LLM acts as a planner (red box) that provides a single step instruction to the actor (green box). In both baselines, the planner and actor are the same, namely Llama 2 and CLIPort respectively. The actor provides action policies, i.e., the actions of the robot. The results of those actions are observed by both actor and reporter. In this baseline, the reporter (blue box) is the vision-language model OpenFlamingo. The reporter provides \textbf{captions} that report on the observation state (``an orange block in the orange area, an orange block outside of the orange area'') and an action \& success state (``The last instruction "Pick up the orange block and place it on the orange area" is executed successfully''), which are both sent back to the planner as explicit language-based feedback to produce the next step.}
\label{fig:architecture1}
\centering
\includegraphics[width=\textwidth]{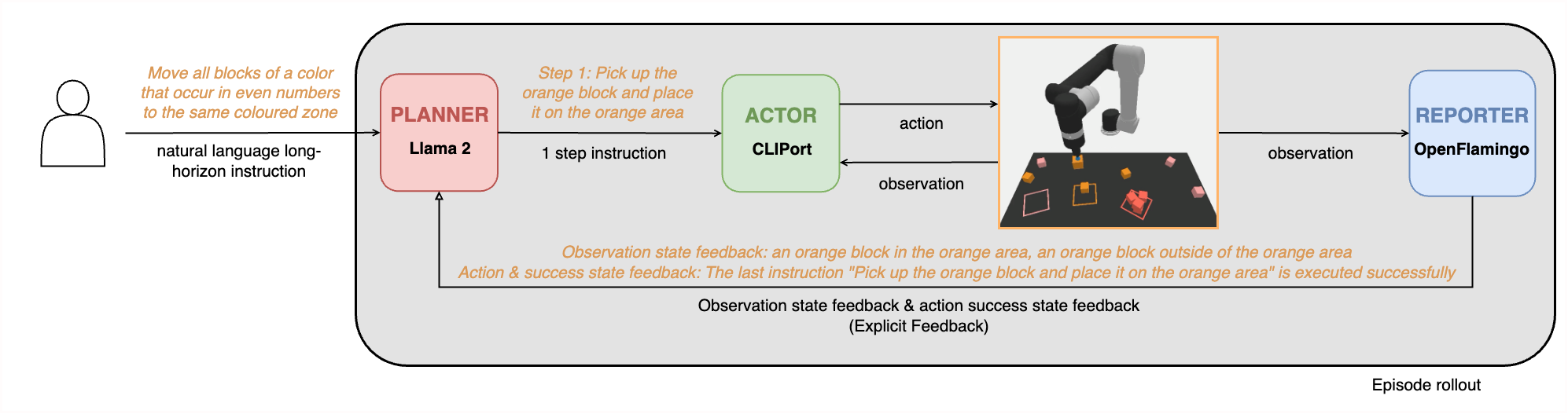}
\end{figure*}

\begin{figure*}[]
\caption{\textbf{Implicit feedback: Learnable interface.} This baseline has the same planner (red box) and actor (green box) architecture as the caption-based baseline in Fig. \ref{fig:architecture1}.  The difference is that in this baseline the reporter (blue box) is a \textbf{learnable interface} (as described in Sec. \ref{sec:learnable}). It provides the translated visual embedding as implicit feedback to the LLM to produce the next step.}
\label{fig:architecture2}
\centering
\includegraphics[width=\textwidth]{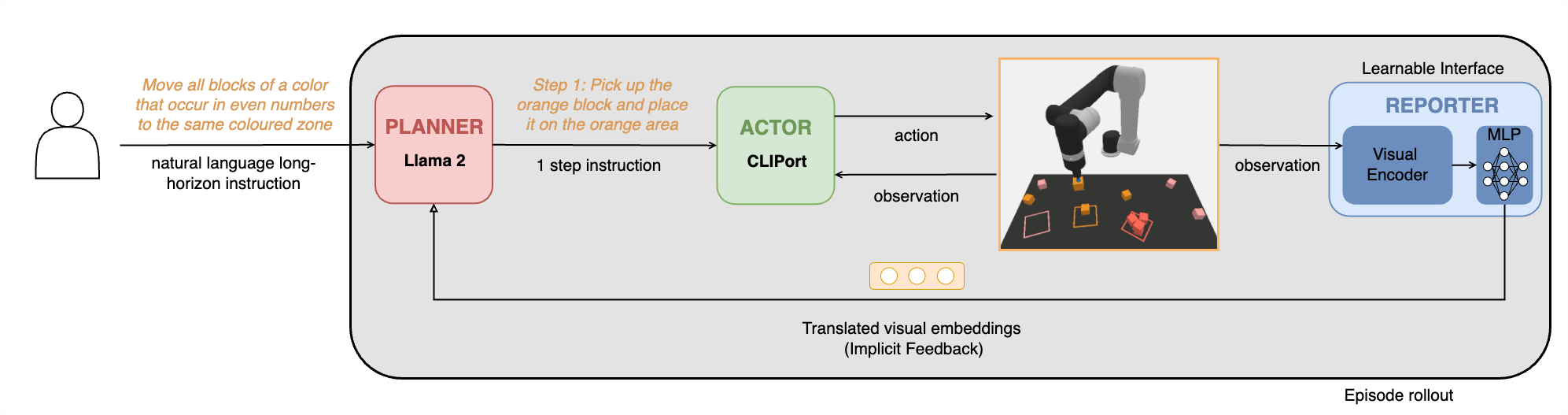}
\end{figure*}

As LLMs show more and more impressive emergent abilities in
various fields, it has been a mainstream method to use
LLMs as the planner for a robot's execution.  However,
most prior work combining LLMs and robots assumes that the
planning information flows unidirectionally from LLMs to
robots, neglecting the role of feedback from the environment
and the robot in LLM planning.  Therefore, how to
incorporate real-time visual observation feedback into the
LLM's input is an under-explored problem.  This modality gap
is especially severe for long-horizon robotic tasks because
an execution error in each of the robot's steps can affect all the
following steps.


To solve the above modality bridging problem, we propose two methods to translate the visual observation into feedback that the LLM can understand for its closed-loop planning. 
Both of these methods will serve as baselines for our proposed LoHoRavens benchmark.
We use the Planner-Actor-Reporter paradigm introduced by~\cite{dasgupta2022collaborating} to unify our two baselines.
The feedback generation models of the two baselines are working as the Reporter module.

\subsection{Explicit feedback: Caption generation}
Inner Monologue~\cite{pmlr-v205-huang23c} demonstrated that human-provided language feedback can significantly improve high-level instruction completion on robotic manipulation tasks. 
But human-written language feedback is too expensive to scale.
We therefore explore a caption generation based model as an automatic way to generate language feedback without training.

As shown in Fig. \ref{fig:architecture1}, we use Llama 2 13B~\cite{touvron2023llama} and the trained pick-and-place CLIPort primitive as the Planner and Actor, respectively.
For the Reporter,
we use  VLM OpenFlamingo~\cite{awadalla2023openflamingo, anas_awadalla_2023_7733589, Alayrac2022FlamingoAV} with few-shot prompting.
Theoretically, any type of feedback from the environment and
the robot can be considered to inform the LLM planner as
long as it can be stated verbally. However, considering the
LoHoRavens simulated environment and the VLMs we use, we
just prompt the VLMs to generate the following two types of feedback. 

\enote{hs}{i would prefer that we write REPORTER PLANNER
  OBSERVER (and also ACTOR?)}

\paragraph{Observation state feedback}
Besides the human instruction at the beginning, the Planner needs to have the information about the objects on the table for the planning. 
Furthermore, if the states of the objects change, the VLM Reporter should describe the changes to the LLM Planner.
\paragraph{Action and success state feedback}
The robot Actor may fail to complete the instruction given
by the LLM Planner.
This kind of success state information
(or rather failure information)
should be conveyed to the Planner.
The VLM Reporter will indicate in its description whether the last instruction is executed successfully or not.

For each seen task in LoHoRavens, we create 10-shot examples
for both LLM prompts and VLM prompts. We use the same few-shot example prompts for the unseen tasks.
When a step's action has executed, there will be a top-down RGB image rendered by the simulator. 
The VLM as the Reporter module will generate the caption feedback based on the current image or the whole image history.
This caption feedback is sent to the LLM for its next-step planning.
The Planner-Actor-Reporter closed-loop process will be
iteratively executed until the high-level goal is achieved
or the maximum number of trial steps has been exceeded.

\subsection{Implicit feedback: Learnable interface}
\label{sec:learnable}
Explicitly converting an image to language captions
is
is straightforward and simple. However,
it typically causes information loss~\cite{zhang2023multimodal, li2023videochat} and exaggerates bias present in training data~\cite{hendricks2018women}.
On the other hand, training an end-to-end multimodal LLM would be too expensive.
Thus another common solution used in many vision-language
models is to use a learnable interface such as a
projection-based interface~\cite{liu2023llava} or a group of
learnable query tokens~\cite{dai2023instructblip} to connect
vision and language modalities while freezing parameters of
the LLM and the visual encoder. This is our second baseline
approach.


We use LLaVA~\cite{liu2023llava} for this second baseline.
LLaVA uses the simple projection-based scheme as the learnable interface between the vision model and the pretrained LLM.
As shown in Fig. \ref{fig:architecture2}, the pretrained CLIP visual encoder ViT-L/14~\cite{radford2021learning} encodes the observation image to visual embeddings. 
A single-layer MLP as the learnable interface then translates the visual embeddings to the LLM's token embedding space. 
The LLM will generate the next-step plan conditioned on the language instruction prompts and the translated visual embeddings.
LLaVA uses LLaMA as the LLM.
To unify this architecture into the Planner-Actor-Reporter paradigm, we can regard LLaMA as the Planner, CLIPort as the Actor, the learnable interface single-layer MLP and the CLIP viusal encoder ViT-L/14 constitute the Reporter module.

To fine-tune LLaVA, for each step of the task instances in the train set, we use the oracle program of the simulator to generate the image before the step and the language instruction for the step as the pair of train data.
For the inference process, LLaVA receives the generated
images after each step's execution
(just as the caption generation based model does).
LLaVA then outputs the next-step language instruction to CLIPort for execution. 

\section{Experiments}
In this section, we aim to answer the following two questions:

\emph{(1) Is our proposed LoHoRavens benchmark a challenging benchmark for current popular models?}

\emph{(2) Which method of incorporating the visual
observation feedback to LLMs is better for long-horizon
robotic manipulation tasks: implicit or explicit?}

\subsection{Experimental settings}
There are two baselines in our experiments: explicit caption
based model and implicit learnable interface based model. 
For the caption based model, we can further compare the
effects of each module of Planner, Actor, and
Reporter. Except for the {\tt CLIPort (oracle)} model, all
the other models use the same pick-and-place primitive
Actor trained on three sets
(one for each of the three primitives)
of 20,000 demonstrations by multi-task learning.

{\tt CLIPort (oracle)}
refers to using CLIPort as the actor model (without using a
planner or a reporter).
It
is a multi-task policy trained on all the training data of
the seen tasks.
Because the vanilla CLIPort does not know when to stop execution, following Inner Monologue and CaP, we use an oracle termination variant that uses the oracle information from the simulator to detect the success state and stop the execution process.
{\tt CLIPort + Llama 2}~\footnote{We use the Llama 2 13B version.} is the model combining Actor and Planner. 
{\tt CLIPort + Llama 2 + OpenFlamingo}~\footnote{We use the OpenFlamingo-9B-vitl-mpt7b version.} is the model combining Actor, Planner, and Reporter.
Both Llama 2 and OpenFlamingo use 10-shot prompts for inference.
For the learnable interface model, {\tt LLaVA}~\footnote{We
fine-tune the LLaVA 13B version.} serves as the Reporter and
Planner modules. As mentioned before (end of Sec.\ \ref{sec:learnable}), we fine-tune it on our generated training data consisting of pairs of simulator rendered images and corresponding next-step language instruction.

\subsection{Experimental results}

Table~\ref{tab:lohoravens} gives experimental results.  The
results show that the performance of all models is quite
poor on almost all tasks, which indicates LoHoRavens is a
quite challenging benchmark for current popular LLMs and
VLMs.  We find that all models perform better on  tasks
requiring reasoning about only one aspect/attribute (e.g., tasks B and
D) than on tasks involving several (e.g., 
size and color in task C, arithmetics and color in task E).
Combining several types of reasoning capabilities is
apparently challenging for the models.


Comparing the results of {\tt CLIPort (oracle)}, {\tt
  CLIPort + Llama 2}, and {\tt CLIPort + Llama 2 +
  OpenFlamingo}, we find that both LLM and VLM usually
improve the single CLIPort model. The VLM is especially
helpful when execution errors are likely to occur, such as
the stacking tasks G and H where an error in one step such
as dropping a block may easily affect previously stacked
blocks.  However, we can also notice in some tasks requiring
reference capability (like tasks E and K) that the LLM
brings negative effects. We conjecture that this is because
the LLM cannot give the precise description to indicate
which block should be manipulated when there are several
objects of the same size and color.

We also see that the learnable interface
based model can outperform
the caption based model
in tasks where the observations are complex in
the sense that they are difficult to describe in
language. For example, in task B, there are too many objects
of the same size and similar color to recognize. In task D,
some objects are unobservable if other objects are stacked
on them. This may be the reason that the VLM fails to give
an accurate description for the image in these
situations. But {\tt LLaVA} has been trained on the images of the
LoHoRavens environment, so it would be more competent to deal with these complicated images than the caption model without training.

Furthermore, when transferring to the unseen tasks, both the performance of the caption-based model and the learnable interface-based model drops noticeably. However, we find the caption-based model is more robust to the unseen tasks than the learnable interface model. We think the reason is that the training-free caption-based model is less affected by whether the task is new or not.

\enote{hs}{maybe add seomthign like: in summary, the
  learnable interface is more expressive, but the caption
  geneation is more robust. eithre add it here or in the
  conclusino or in both}

Our findings suggest that LoHoRavens can guide research on several of the main
challenges in this area: (i) how to design models with
reasoning abilities, (ii) how best to provide feedback for
planner/actor, (iii) how to represent
information that is difficult to describe in language, (iv)
how best to achieve good generalization for unseen tasks.

\section{RELATED WORK}

\enote{hs}{in this section it is often not clear 
  what the relationship of the cited work is to this work}
\subsection{Language-conditioned robotic manipulation benchmark}
The interest in training language-conditioned models for robot manipulation has been growing in recent years thanks to the significant advancements in language processing techniques. 
As a result, many researchers proposed different language-conditioned robotic manipulation datasets and benchmarks. 
RLBench~\cite{james2020rlbench}, Ravens~\cite{zeng2021transporter, shridhar2022cliport}, Robosuite~\cite{robosuite2020} introduce various manipulation tasks in the household or the tabletop environment household tasks with their corresponding natural language instructions. 
VIMA-Bench~\cite{jiang2023vima} is a robot manipulation learning benchmark supporting multimodal-prompting tasks.
VLMbench~\cite{zheng2022vlmbench} contains multiple 3D manipulation tasks with compositional language instructions and categorizes manipulation tasks into various meta manipulation actions by constraints for the first time.
RM-PRT benchmark~\cite{ren2023rm} designs four progressive reasoning tasks and integrates the instruction parsing capabilities of LLMs. 
ARNOLD benchmark~\cite{gong2023arnold} addresses the challenge of understanding continuous object states in complex tasks, emphasizing the need for language-grounded learning with continuous goals. 
LEMMA~\cite{gong2023lemma} introduces a benchmark for Language-Conditioned Multi-robot Manipulation, specifically focusing on collaboration between robots, task allocation, and handling strong temporal dependencies.

However, all of these benchmarks do not focus on long-horizon tasks.

Inner Monologue~\cite{pmlr-v205-huang23c}, CoP~\cite{codeaspolicies2022}, and Language-Table~\cite{lynch2022language-tabel} build datasets for long-horizon language-conditioned manipulation tasks, but all of their long-horizon datasets are not open-sourced. 
A more specific, yet important benchmark is introduced by \cite{zhao2022opend} with OpenD, a benchmark for language-driven door and drawer opening. Their system employs a multistep planner integrating deep neural networks and rule-based controllers, showcasing promising zero-shot performance but highlighting challenges in language understanding, spatial reasoning, and long-term manipulation. These challenges are those that LoHoRavens tries to explicate in a dedicated benchmark that goes beyond the presented works.

The most similar work to our proposed LoHoRavens is
CALVIN~\cite{mees2022calvin}, which is also a long-horizon language-conditioned manipulation benchmark. However, CALVIN provides step-by-step instructions to help the robot complete the high-level goal, meaning the robot does not need to reason and plan for each step by itself. 
Our LoHoRavens only gives the high-level instruction and tests the robot's long-horizon reasoning and planning capabilities.

\subsection{Foundation models for robot learning}
The emergent abilities of LLMs such as ChatGPT~\cite{ouyang2022training}, GPT-4~\cite{OpenAI2023GPT4TR}, PaLM~\cite{chowdhery2022palm}, LLaMA~\cite{touvron2023llama} has brought big breakthroughs to many fields, as well as the robotics field due to its rich knowledge and strong reasoning capabilities.
At the same time, VLMs also have a remarkable progress, such as CLIP~\cite{radford2021learning}, BLIP-2~\cite{li2023blip2}, InstructBLIP~\cite{dai2023instructblip}, Flamingo~\cite{Alayrac2022FlamingoAV}, LLaVA~\cite{liu2023llava}, MiniGPT-4~\cite{zhu2023minigpt}, whose capabilities can be extended to robotic closed-loop control to enable new levels of generalization.
There are also (multimodal) LLMs such as SayCan~\cite{saycan2022arxiv}, PaLM-E~\cite{driess2023palme}, RT-1~\cite{brohan2022rt1}, RT-2~\cite{brohan2023rt} which are especially designed for robot learning. With them, robots show more and more impressive capabilities in various scenarios.
Our work uses these LLMs and VLMs to explore a solution to the very challenging long-horizon language-conditioned tasks.


\newpage

\devour{
\section{CONCLUSIONS}
We propose a long-horizon language-conditioned tabletop manipulation benchmark LoHoRavens, and two baselines for the challenging benchmark.
Experimental results show that the learnable interface baseline is more expressive, but the caption generation baseline is more robust.
For the future work, LoHoRavens can be extended to include more tasks and more reasoning capabilities.
Further strategies or methods such as code as policies should be explored to improve the long-horizon reasoning for robots.
}





\section*{ACKNOWLEDGMENT}
We would like to thank helpful discussions from Ruotong Liao and Gengyuan Zhang at LMU Munich.
This work was partially funded by the European Research Council (grant \#740516).
This work was also supported by the DAAD programme Konrad Zuse Schools of Excellence in Artificial Intelligence, sponsored by the Federal Ministry of Education and Research.


\bibliographystyle{IEEEtran}
\bibliography{IEEEabrv,mybibfile}



\end{document}